\documentclass[11pt,a4paper]{article}

\usepackage{fullpage}
\newcommand{\say}[1]{``#1''}
\usepackage[round]{natbib}
\usepackage{graphicx}

\DeclareGraphicsExtensions{.png,.jpg}

\newcommand{\keywords}[1]{\small{\textbf{Keywords:} {#1}}}

\usepackage{graphicx}
\usepackage{subfig}

\usepackage{authblk}

\begin{document}

\title{\huge\bfseries Eye-Movement behavior identification for AD diagnosis}

\author[1,2]{Juan Biondi \footnote{Corresponding Author. \\ Email: \textit{juan.biondi@uns.edu.ar}}}
\author[1]{Gerardo Fernandez} 
\author[2,1]{Silvia Castro} 
\author[1,3]{Osvaldo Agamennoni}

\affil[1]{Laboratorio de Desarrollo en Neurociencia Cognitiva, Instituto de Investigaciones en Ingenier{\'i}a El{\'e}ctrica (IIIE), Departamento de Ingenier\'ia El\'ectrica y de Computadoras (DIEC), Universidad Nacional del Sur (UNS) - CONICET, Bah\'ia Blanca, Argentina}
\affil[2]{Laboratorio de Visualizaci\'on y Computaci\'on Gr\'afica (VyGLab), Departamento de Ciencias e Ingenier\'ia de la Computaci\'on (DCIC), Universidad Nacional del Sur (UNS), Bah\'ia Blanca, Argentina}
\affil[3]{Comisi{\'o}n de Investigaciones Cient{\'i}ficas de la Provincia de Buenos Aires (CIC), Argentina}

\date{}

\maketitle

\vspace{-1cm}

\begin{abstract}
In the present work, we develop a deep-learning approach to differentiate between the eye-movement's behavior of people with neurodegenerative diseases over healthy control subjects, from reading. 
The subjects with and without Alzheimer's disease read well-defined and previously validated sentences including high-, low-predictable sentences, and proverbs. 
From these eye-tracking data we derive trial-wise information consisting of descriptors that capture the reading behavior of the subjects. 
With this information we train a set of \textit{denoising sparse-autoencoders} and build a deep neural network using the trained \textit{autoencoders} and a \textit{softmax} classifier that allows identifying patients with Alzheimer’s disease with $89.78\%$ of accuracy. 
Our results are very encouraging and show that these models promise to be helpful to understand the dynamics of the eye movement behavior and its relation with the underlying neuropsychological processes.
\end{abstract}

\keywords{Eye-tracking, Deep-learning, Alzheimer's Disease}

\section{Introduction}
Alzheimer's disease (\textit{AD}) is a nonreversible neurodegenerative disease characterized by progressive impairment of cognitive and memory functions that develops over a period of years being the most prevalent cause of dementia in elderly subjects.
Initially, people experience memory loss and confusion, which may be mistaken for the kinds of memory changes that are sometimes associated with normal aging \cite{waldemar2007recommendations}.
The subtle changes in behavior and response of the early manifestation of this disease make it difficult to diagnose by using the classical neuropsychological tests such as the Mini-Mental State Examination.
The use of more advanced diagnosis tools such as MRI and PET results is critical for its early diagnosis. 
Since \textit{AD} is nonreversible, its early treatment can improve the patient's life delaying the full manifestation of the disease.
In the last years, the study of the eye movement, known as \textit{eye-tracking}, during reading, has proved to help performing this task \citep{fernandez2015patients, fernandezpatients, fernandez2015diagnosis, fernandez2013eye}.

Reading is a cognitive activity that has received considerable attention of researchers to evaluate human cognitive performance. 
This requires the integration of several central cognitive subsystems, from attention and oculomotor control to word identification and language comprehension. 
Eye movements show a reproducible pattern during normal reading. 
Each eye movement ends up in a fixation point, which allows the brain to process the incoming information and to program the following saccade. 
Different neuropsychiatric pathologies produce abnormalities in eye movements and disturbances in reading, having each of them a particular pattern that can be registered and measured \citep{fernandez2016word, fernandez2016contextual, holzman1974eye, iacono1992smooth, riby2009looking, kellough2008time}.
Eye movements can be classified into three groups:
\begin{itemize}
    \item Movement for maintaining the image on the fovea (area of the retina with acuity vision), compensating head or object movements;
    \item Movements for shifting the eyes, when the attention changes from one object to another. There are subtypes of shifting movements: saccades (looking for a new center of visual attention), monitoring and vergence (slower than saccades and are responsible for carrying the image of interest to both foveae, allowing stereoscopic vision);
    \item Movements of binocular fixation that also prevent fading of the image. These movements have three variations: tremor, drift, and microsaccade.
\end{itemize}
Saccades are rapid big eye movements particularly important from the cognitive point of view since cognitive processes have a direct influence on such movements. 
Each saccade has its direction. 
People, depending on language, read from left to right and most of the saccadic eye movements are oriented accordingly. 
These normal reading movements are called forward saccades. 
Reading movements going from right to left are called regressions. 
The saccade movement alternates with a fixation made when the eyes are directed to a particular target (See \citep{rayner1998eye} for a review).
As shown in \citep{fernandez2013eye}, patients with early Alzheimer disease show alterations during the execution of tasks, such as reading, and these alterations can be related to an impairment in their working memory \citep{fernandez2014registering, fernandezpatients}. 
In fact, it has been shown that through this differentiation in the eye-movements, it is possible to infer a diagnosis \citep{fernandez2015diagnosis}.

The use of computer-aided diagnosis is a key challenge since the growth of computational power permits the creation of more complex models. 
These models can be used to create biomarkers that help in disease identification.
Since the popularization of \textit{deep-learning} neural networks \cite{schmidhuber2015deep, deng2014deep}, many efforts have been made in their use in the field of medicine.
This technique is commonly used in conjunction with imaging diagnoses such as PET or MRI mainly because the feature representation that this technology provides may help even when data is incomplete \cite{li2014deep}.
Specifically, there were advances in the detection and pattern differentiation of the physical brain alterations that neurodegenerative diseases produce, such as \textit{AD} and mild cognitive impairment (MCI) \cite{suk2013deep, suk2014hierarchical}.
Even there were advances in its early diagnostic \cite{liu2014early}.
The problem is that, when a brain physical alteration is observable, the damages made to the brain are irreversible (even though the disease is in an early stage) and may cause deterioration in the quality of life of the patient.
The eye-tracking technique allows us to find subtler changes that were made by the brain to alleviate small memory deficits in the patient.
These changes are not noticed by the patient but small changes in the way they read our set of sentences can be found with the technique presented in this paper.

In this work, we use a deep-learning neural network trained on reading information extracted from controls and patients with probable \textit{AD} in order to identify the patterns made by them in the reading process and later cluster them in their respective groups.
Throughout this work, we may use \textit{AD} patients and patients with probable \textit{AD} in an interchangeable manner because of the nature of the \textit{AD} diagnosis.
The hypothesis was that using deep-learning in the feature identification of the key characteristics of the patient's eye behavior during reading sentences may lead to a correct classification that can be used to infer a diagnosis.
Using this type of technology may improve the results obtained in \citep{fernandez2015diagnosis} since it provides a smaller granularity in the detection of the disease and consequently, a better performance.
Additionally, this technology allows us to improve the effectiveness of the classification as we collect more ground truth subjects.

\section{Methods}

\subsection{Ethics Statement}
The investigation adhered to the principles of the Declaration of Helsinki and was approved by the Institutional Bioethics Committee of the Hospital Municipal de Agudos (Bah{\'i}a Blanca, Buenos Aires, Argentina). 
All patients and their caregivers, and all control subjects signed an informed consent prior to their inclusion in the study.

\subsection{Participants}
Twenty six patients (mean age $69$ years, $SD=7.3$ years) with the diagnosis of probable \textit{AD} were recruited at Hospital Municipal of Bah{\'i}a Blanca (Buenos Aires, Argentina). 
The clinical criteria to diagnose \textit{AD} at its early stages remains under debate \cite{mckhann1984clinical}. 
In the present work, the diagnosis was based on the criteria for dementia outlined in the Diagnostic and Statistical Manual of Mental Disorders (DSM-IV). 
All \textit{AD} patients underwent a detailed clinical history revision, physical/neurological examination and thyroid function test. 
They all presented an APO E3E4 Genotype. 
Magnetic resonance images were obtained from twelve patients and computerized tomography scans from the other. 
All the patients underwent biochemical analysis to discard other common pathologies (hemoglobin, full blood count, erythrocyte sedimentation rate, urea and electrolytes, blood glucose). 
All this data provided a more precise diagnosis of \textit{AD}. 
Patients were excluded if: (1) they suffered from any medical conditions that could account for, or interfere with, their cognitive decline; (2) had evidence of vascular lesions in computed tomography or FMRI; (3) had evidence for an Axis I diagnosis (e.g. major depression or drug abuse) as defined by the DSM-IV. 
To be eligible for the study, patients had to have at least one caregiver providing regular care and support. 
Patients taking cholinesterase inhibitors (ChE-I) were not included. None of the subjects were taking hypnotics, sedative drugs or major tranquilizers. 
The control group consisted of $43$ elderly adults (mean age $71$ years, $SD=6.1$ years), with no known neurological or psychiatric disease according to their medical records, and no evidence of cognitive decline or impairment in daily activities. 
A one-way ANOVA showed no significant differences between the ages of \textit{AD} and \textit{Control} individuals. 
Those participants diagnosed of suffering from Ophthalmologic disease such as glaucoma, visually significant cataract or macular degeneration as well as visual acuity less than 20/20 were excluded from the study.

The mean scores of \textit{Controls} and \textit{AD} patients in the Mini-Mental State Examination (MMSE) \cite{folstein1975mini} were $27.8$ ($SD=1.0$) and $24.2$ ($SD=0.8$), respectively, the latter suggesting early mental impairment. 
A one-way ANOVA evidenced significant differences between MMSE in \textit{AD} patients and \textit{Controls} ($p<0.001$). The mean score of \textit{AD} patients in the Adenbrook's Cognitive Examination - Revised (ACE-R) \cite{mioshi2006addenbrooke} was $84.4$, ($SD=1.1$), the cut-off being of $86$.
The mean school education trajectories in \textit{AD} patients and \textit{Controls} were $15.2$ ($SD=1.3$) years and $15.1$ ($SD=1.0$) years, respectively. 
A one-way ANOVA showed no significant differences between the education of \textit{AD} and \textit{Control} individuals. 

\subsection{Apparatus and eye movement data}
Single sentences were presented on the center line of a 20-inch LCD Monitor ($1024$\textit{x}$768$ pixels resolution; font: regular; New Courier; 12 points, $0.2^{\circ}$ in height). Participants sat at a distance of 60 cm from the monitor. Head movements were minimized using a chin rest. Correction for the 60 cm viewing distance was performed by using the \textit{Eyelink1000} corneal reflection system, which assessed changes in gaze position by measuring both the reflection of an infrared illuminator on the cornea and the pupil size, by means of a video camera sensitive to light in the infrared spectrum.

Eye movements were recorded with an \textit{EyeLink1000 Desktop Mount} (SR Research) eye tracker, with a sampling rate of 1000 Hz and an eye position resolution of a 20-s arc. 
All recordings and calibration were binocular. 
Eye movement data from $69$ participants reading $184$ sentences resulted in a total of $48716$ fixations - $13002$ for \textit{Control} and $35714$ for the people with \textit{AD}. 
This data was cleaned from blinks and track losses. 
Prior to removing the analysis fixations shorter than $51ms$ and longer than $750ms$, and fixations on the first and last word of each sentence (see \citep{kliegl2006tracking} for a description of the analytic procedure), we measured, for each patient, the elapsed time between the instant when sentences were first presented, and the instant when participants looked at the final spot: mean reading time in high-predictable sentences was $3495 ms$ vs. $5635 ms$ (Controls and \textit{AD}) and $4828 ms$ vs. $6881 ms$ (controls and \textit{AD}) in low predictable-sentences.

\subsection{Procedures}
Participant's gaze was calibrated with a standard 13-point grid for both eyes. 
After validating the calibration, a \textit{trial} began with the appearance of a fixation point on the position where the first letter of the sentence was to be presented. 
As soon as both eyes were detected within a radius equal to $1^{\circ}$ from the fixation spot, the sentence was presented. 
After reading it, participants looked at a dot in the lower right corner of the screen; when the gaze was detected on the final spot, the \textit{trial} ended. Occasionally, external factors such as minor movements and slippages of the head gear could cause small drifts. To avoid them, we performed a drift correction before the presentation of each spot.

To assess whether subjects comprehended the texts, they were presented with a three alternative multiple-choice question about the sentence in progress in $20\%$ of the sentence \textit{trials}. Participants answered the questions by moving a mouse and choosing the response with a mouse click. Overall mean accuracy was $95\%$ ($SD=3.2\%$) in \textit{Control} and $91\%$ ($SD=5.4\%$) in \textit{AD}. A one-way ANOVA showed no significant differences between comprehension of the answers in \textit{Controls} and in \textit{AD} patients. The latter were only marginally less accurate than \textit{Control} subjects, probably because they were in an early stage of the pathology, as indicated by the MMSE and ACE-R values. Once the comprehension test ended, the next \textit{trial} started with the presentation of the fixation spot. An extra calibration was done after 15 sentences or if the eye tracker did not detect the eye at the initial fixation point within 2 s.

\subsection{Sentence corpus}
The sentence corpus was composed of short sentences of a line with $75$ low predictable sentences, $45$ high-predictable sentences and $64$ proverbs (e.g., \say{Maria is always laughing and in a good mood}, \say{It is worthwhile to think before talking} and \say{A bird in the hand is worth two in the bush})\cite{fernandez2014eye}. 
All the sentences comprised a well-balanced number of content and function words and had similar grammatical structure. 

\subsubsection{Word and Sentence Lengths} 
Sentences ranged from a minimum of $5$ words to a maximum of $14$ words. 
Mean sentence length was $8.1$ ($SD=1.4$) words for low predictability sentences, $7.6$ words ($SD=1.5$) for high predictability sentences and $7.3$ words ($SD=1.9$) for proverbs. 
Words ranged from $1$ to $14$ letters. 
Mean word length was $4.6$, $4.1$ and $4.0$ ($SD=2.5$, $SD=2.3$ and $SD=2.0$) for low-, high-predictable sentences and proverbs, respectively.

\subsubsection{Word Frequencies}
We used the Spanish Lexical Lexesp corpus \cite{sebastian2000lexesp} for assigning a frequency to each word of the sentence corpus. Word frequencies ranged from 1 to 264721 per million, so we transformed it to $log_{10}(frequency)$. Mean $log_{10}(frequency)$ was 3.4 ($SD=1.3$) for low predictability sentences, 3.4 ($SD=1.5$) for high predictability sentences and $3.47$ ($SD=1.36$) for proverbs.

\subsubsection{Word Predictability}
It was measured in an independent experiment with $18$ researchers of the Electrical Engineering and Computer Science Department of Universidad Nacional del Sur. 
We used an incremental cloze task procedure in which participants had to guess the next word given only the prior words of the sentence. 
Participants guessed the first word of the unknown sentence and entered it via the keyboard. 
In return, the computer presented the first word of the original sentence on the screen. 
Responding to this, participants entered their guess for the second word and so on, until a period indicated the end of the sentence. 
Correct words stayed on the screen. 
Participants were between 31 and 62 years old and did not participate in the reading experiment. 
Academic background of the reading experiment group and the cloze task group was similar. 
Word predictabilities ranged from $0$ to $1$ with a mean of $0.38$ ($SD=0.36$). 
The average predictability measured from the cloze task was transformed using a $logit$ function $0.5*ln(pred/(1 - pred))$; predictabilities of zero were replaced with $1/(2*18) = - 2.55$ and those among  the five perfectly predicted words with $(2*18 - 1)/(2*18) = +2.55$, where 18 represents the number of complete predictability protocols. 
Mean $logit$ predictability was $- 0.9$ ($SD=0.9$) for low predictability sentences, $0.0$ ($SD=1.29$) for high predictability sentences and $0.08$ ($SD=1.23$) for proverbs.

As in other languages, we find strong correlations in Spanish between word length, word frequency, and word predictability. 
Long words are of low frequency ($r = -0.80$ and $r = -0.75$ in low and in high predictability sentences, respectively). 
Frequent words are highly predictable ($r = 0.47$ and $r = 0.37$ in low and in high predictability sentences, respectively), and highly predictable words tend to be short words ($r = -0.47$ and $r = -0.38$ in low and in high predictability sentences, respectively).

\subsection{Used information}
The information used for this work was a \textit{trial-wise} compaction of the original data where we keep descriptors of the mean reading behavior of the subjects in each read sentence.
We measured the saccade amplitude, fixation duration and duration of the fixation on a single word of the subject during the reading of each sentence but only kept the mean and the standard deviation.
In addition, we measured the total number of fixations and classified them by first pass fixations, refixations, unique fixations and total fixations:
\begin{itemize}
    \item First pass fixations: The first fixation on a specific word of the sentence.
    \item Unique fixations: Fixations that occur once in a word that was skipped in the first pass.
    \item Multiple fixations: Multiple fixations on a word in the first pass.
    \item Refixations: Fixations that take place once a word that already has a \textit{first pass fixation} or a \textit{unique fixation} implying a regression.
\end{itemize}

Categorical data as the diagnosis (used as training labels) was replaced by numerical values in order to unify the data types and improve during the classification process.
An integer with two possible values was used for the diagnosis information construction: $0$ for \say{Control} and $1$ for \say{AD}.
The identification and the diagnosis information of the subject were kept apart from the data.
A detail of the variables used as input for the model construction is shown in table \ref{tab:varnames}.
Since the tag (\textit{AD} or \textit{Control}) is associated to the patient and not to each sentence, and, since we use a per-trial classification approach, the subject's tag was applied to all the sentences read by him/her.
Following this approach may introduce noise in the classifying stage because, as we use a per-trial classification approach, a \textit{Control} subject could, for example, be distracted during the reading of a specific sentence thus making him read as a non-healthy person.
Anyway, the system should be able to detect and ignore these artifacts because many samples are used in the training stage.

\begin{table}[h]
\begin{center}
\caption{Used variables for the model construction.}{\begin{tabular}{  l  l }
    \textbf{Name} & \textbf{Description} \\ \cline{1-2}
    nw & Number of words in the sentence.\\ \hline
    gaze & Global (sentence) mean of the sum of fixation durations on the same word.\\ \hline
    sd\_gaze & Standard deviation of \textit{gaze}.\\ \hline
    as & Mean saccade amplitude in the sentence.\\ \hline
    sd\_as & Standard deviation of \textit{as}\\ \hline
    ntf & Count of the total number of fixations on the sentence.\\ \hline
    ntm & Count of the number of multifixations on the sentence.\\ \hline
    dfp & Mean duration of the first pass fixations on the sentence.\\ \hline
    sd\_dfp & Standard deviation of \textit{dfp}.\\ \hline
    fpp & Count of the number of first pass fixations on the sentence. \\\hline
    rf & Count of refixations on the sentence.\\ \hline
    nfu & Count of unique fixations on the sentence.\\ \hline
    dfu & Mean duration of unique fixations on the sentence.\\ \hline
    sd\_dfu & Standard deviation of \textit{dfu}.\\ \hline
    \end{tabular}}
\label{tab:varnames}
\end{center}
\end{table}

All the data was previously outlier-checked by establishing a dropout policy in order to use a cleaner dataset. 
The outlier check policy consisted of finding the mean and the $SD$ of each condition group and checking the two groups separately.
All the \textit{trials} where the standard deviation was bigger than two times the standard deviation of the group were considered as outliers and was dropped out, resulting in a $10\%$ samples lost.
The sentence identiﬁcation, order, and type were kept separate from the training information.
This is because the standard deviations in the information for the \textit{AD} patients for proverbs and high predictability sentences, appear to be particularly high after the data is outlier-checked, causing a highly unbalanced dataset.

The resulting dataset consisted on $3235$ \textit{trials} with mean $46.88$ ($SD=10.76$) \textit{trials} per subject.
The dataset was splitted in two groups: one for the training of the network with data of $61$ subjects and other for testing with data of $8$ subjects randomly picked.
Finally, the training dataset consisted on $2922$ \textit{trials} of $61$ subjects - $39$ \textit{Control} and $22$ \textit{AD} - with $47.9$ ($SD=10.47$) mean \textit{trials} each; the testing dataset consisted on $313$ \textit{trials} of $8$ subjects - $4$ \textit{Control} and $4$ \textit{AD} - with $39.12$ ($SD=10.37$) mean \textit{trials} each.
Splitting the data in this way ensures that the network can't infer the condition in other way, avoids over-fitting and ensures that the testing data is totally unknown by the network.

\subsection{Deep learning with \textit{denoising sparse-autoencoders}}
In this work we used \textit{sparse-autoencoders} for the codification stage. %feature identification.
The \textit{sparse-autoencoders} work just as regular \textit{autoencoders}, i.e. there are neural networks under supervised learning with the targets set equal to the input (the identity) but, in the case of \textit{sparse-autoencoders}, an average number of activations per neuron restriction was applied in the hidden layer by penalizing the average number of activations different from the desired (known as sparsity proportion) adding a penalty term to the cost function.
This restriction is introduced so that each neuron specializes on a particular feature. 
The lower the sparsity proportion, the more  specific the feature.
The resulting trained neural network can be thought of as: an encoder, involving the input and the hidden layer, and a decoder, involving the hidden and the output layer.
In this case, we set an activation restriction equal to $10\%$.

In a \textit{denoising-autoencoder}, the idea is to force the hidden layer to discover more robust features and to prevent it from simply learning the identity, by training the \textit{autoencoder} to reconstruct the input from a corrupt version.
The altered version of the input was generated by introducing noise, which was obtained by clamping some of the fields to zero.
The corrupt data was used as the \textit{sparse-autoencoder} input, and the clean (unaltered) data as the target.
Using this type of data corruption mechanism forces the network to learn a way of reconstructing a field based on others.
This, combined with the sparsity restriction, results in more robust features.

The deep-learning neural network was built using two stages of these \textit{denoising sparse-autoencoders}. 
In each stage, we train the \textit{autoencoder} by corrupting the encoded clean data obtained from the previous stage, and providing it to the next as its input.
At the end of the two stages, we set a \textit{softmax} layer as a classifier, training it with uncorrupted data and the corresponding tag.
As we used a \textit{per-trial} classification approach, the patient diagnosis was extended to all the sentences read by him and the classifier was trained with this data as the target.
The \textit{softmax} layer is a non-linear, multiclass generalization of the binary \textit{Logistic Regression}, and its output is the \say{probability} for each class (we quoted the word \say{probability} because it's shape depends on of the regularization used in the training stage, it can be more diffuse or peaky).

\section{Results}
\begin{figure}[h]
\centering
\includegraphics[width=0.6\textwidth]{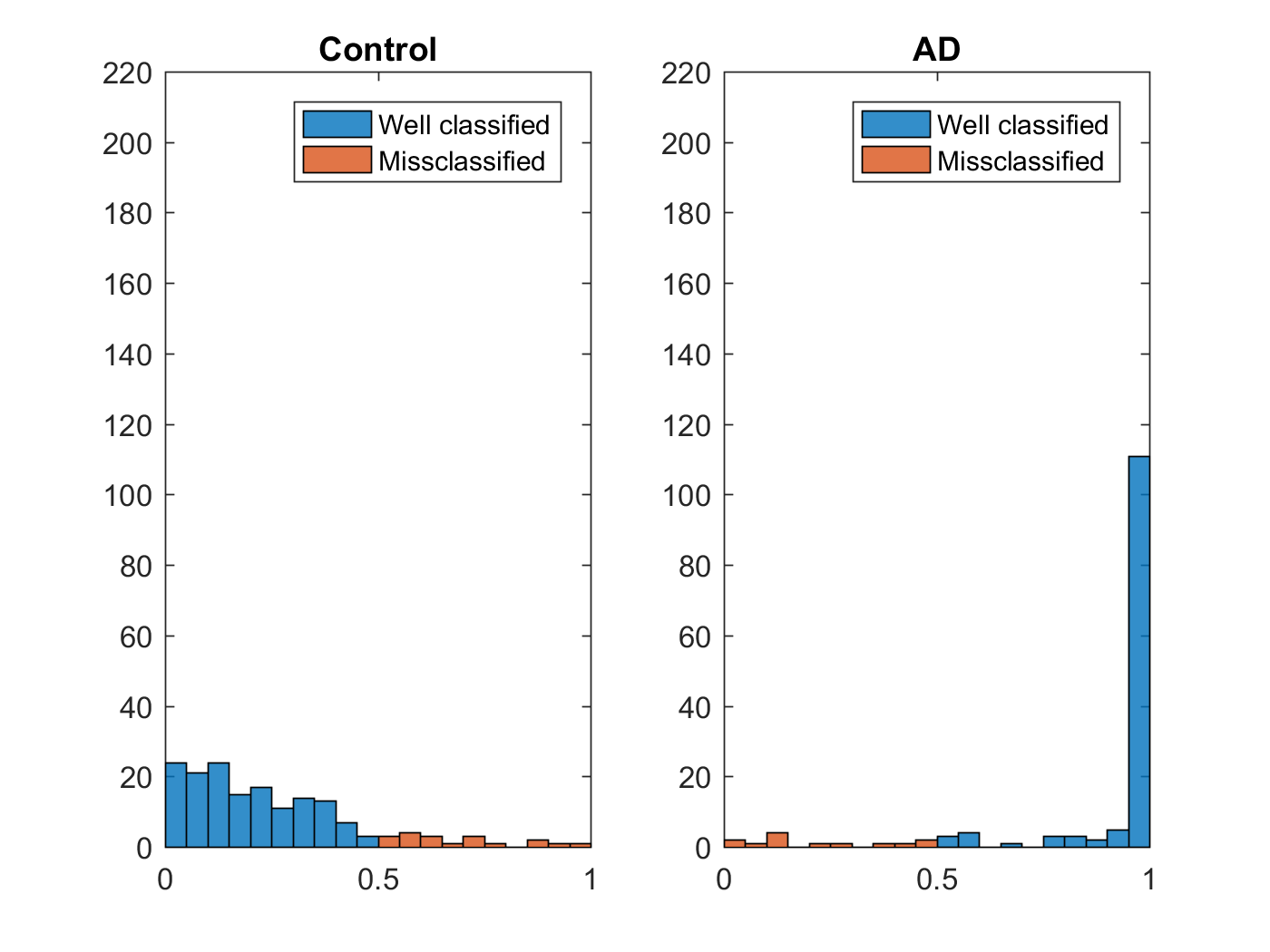}
\caption{Classification results histogram representing number of sentences, split by \say{ground truth} values. Values below 0.5 are considered as classified as \textit{Control}, and higher values are considered as classified as \textit{AD}.}
\label{fig:hist}
\end{figure}

Several configurations were generated by varying the sparsity proportion, the number of units and layers and the shape of the network (same vs. decreasing number of units between layers).
We adopted the one that produced the best results which consisted of two layers of \textit{denoising sparse-autoencoders} with $16$ and $4$ hidden units in the hidden layer each, using a sparsity proportion of $10\%$.
After the training of the network, a series of tests were performed with data not included in the training dataset.
This dataset, as mentioned, was composed of $313$ sentences from $8$ subjects - $4$ \textit{Control} and $4$ \textit{AD} - with $39.12$ ($SD=10.37$) mean \textit{trials} each.
We used a \textit{softmax} layer for the classification trained using the condition translation with $0$ for \textit{Control} subjects and $1$ for \textit{AD} subjects. 
This means that we have a single class \say{AD} and, since the output of the classifier is a real number between $0$ and $1$, the read sentences classified by the network with values close to $0$ have a small \say{probability} of being read by an \textit{AD} patient (i.e. high probability of being read by a \textit{Control} patient) and vice-versa. 
The \say{ground truth} values are known, so we can split the output into groups and observe the number of sentences misclassified by the network.
Based on this, we show the output of the network where values below $0.5$ are considered as classified as \textit{Control}, and higher values are considered as classified as \textit{AD} (see Figure \ref{fig:hist}). 
As can be seen, the output of the network was consistent with the expected values.

\begin{figure}[h!]
\centering
\includegraphics[width=0.6\textwidth]{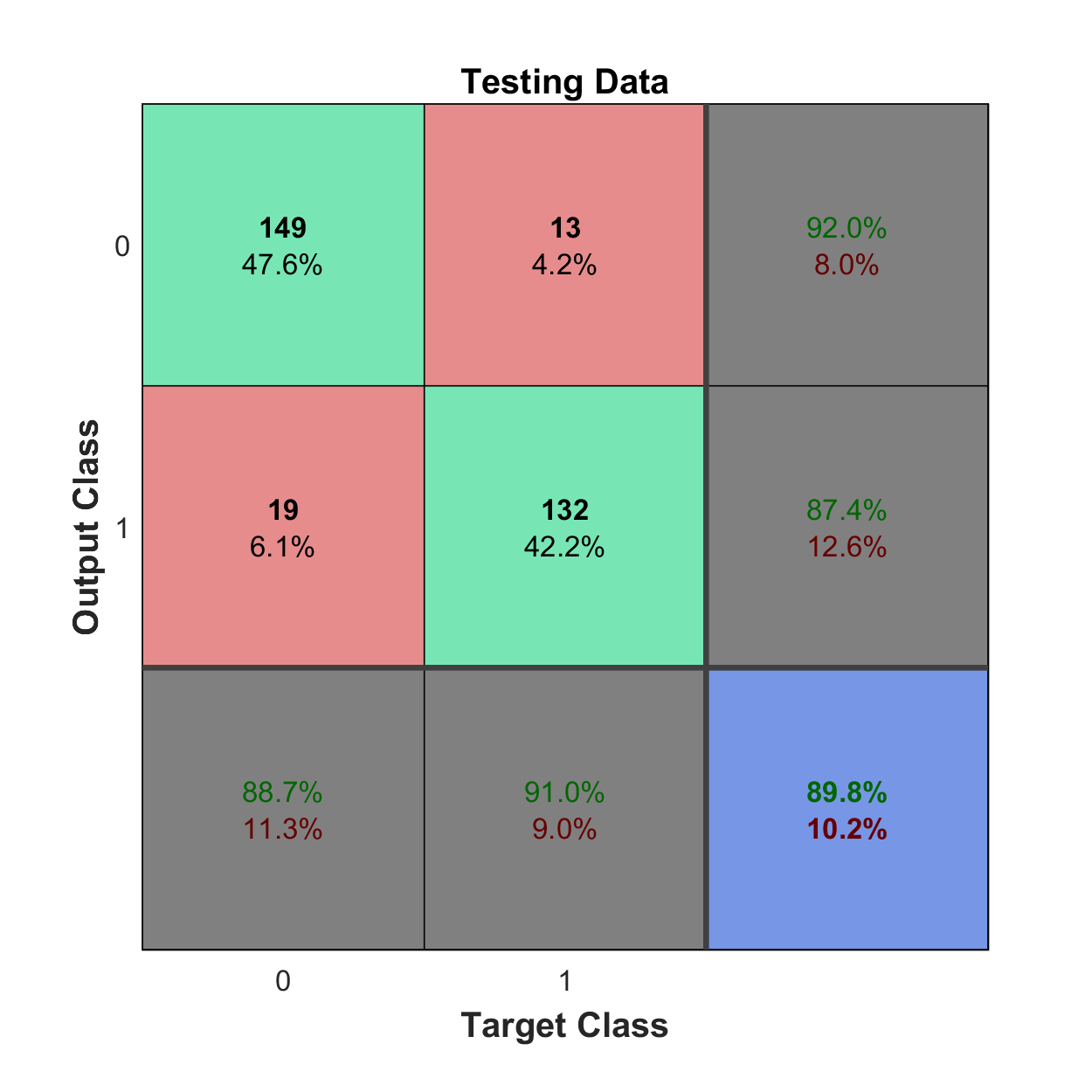}
\caption{Classification results. Values below 0.5 are considered as classified as \textit{Control} (class 0), and higher values are considered as classified as \textit{AD} (class 1).}
\label{fig:confusion}
\end{figure}

Now, we can round the values so we can plot a confusion matrix and approximate the number of misclassified sentences in order to measure the performance of the network.
In the figure \ref{fig:confusion} we can see a confusion matrix of the output.
The \say{X} axis represents the expected output values and the \say{Y} axis the rounded output of the network.
As can be seen, the overall performance of the network was good giving a $89.8\%$ of well-classified sentences.
The performance of the network using sentences read by \textit{Control} patients ($88.7\%$ correctness) approximates to the performance observed using sentences read by \textit{AD} patients ($91.0\%$ correctness).

\begin{figure}[h!]
\centering
\includegraphics[width=0.6\textwidth]{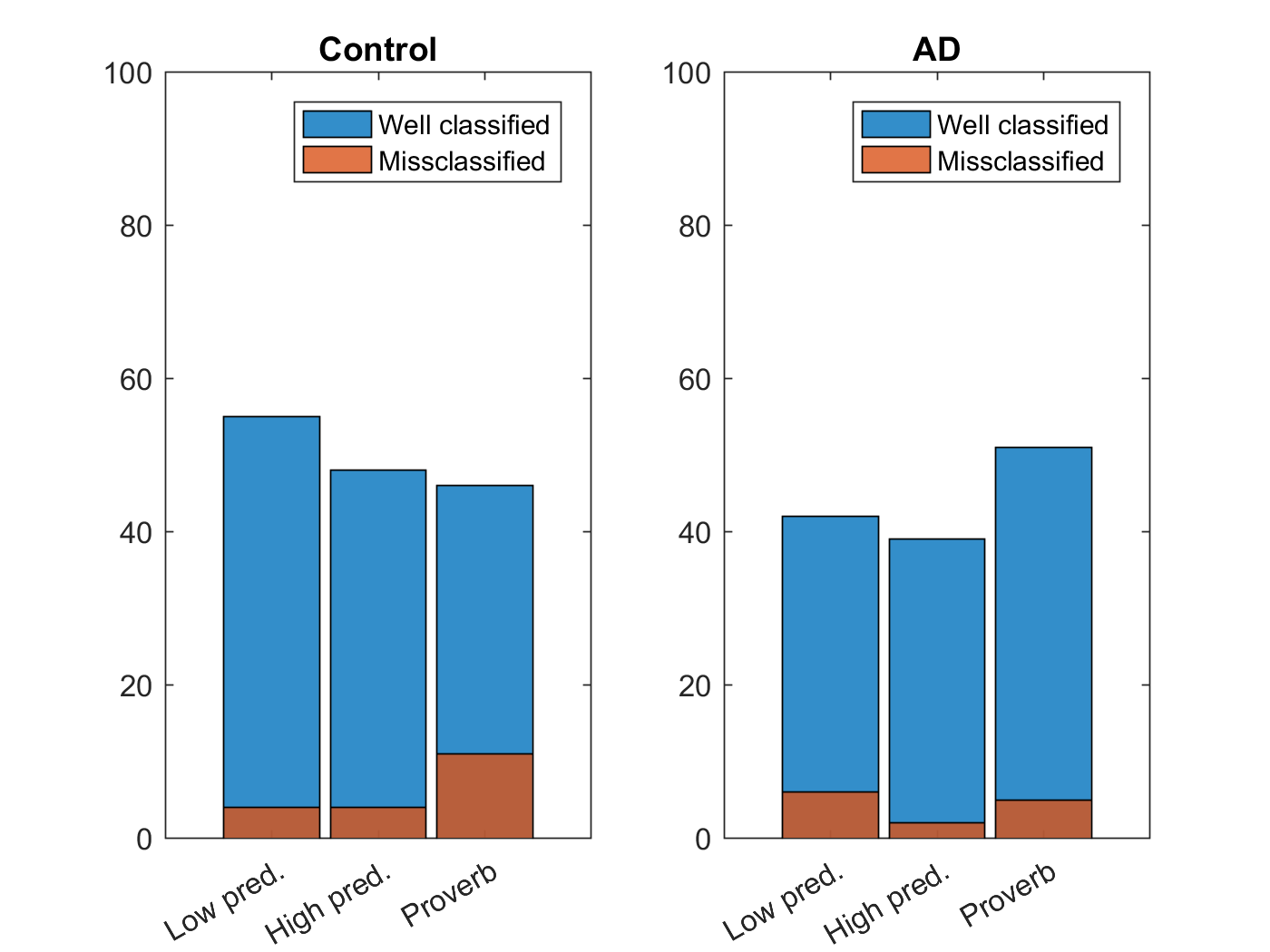}
\caption{Number of misclassified sentences by type, split by \say{ground truth} label.}
\label{fig:type}
\end{figure}

On the other hand, misclassified sentences were not concentrated on a particular type of sentence as can be seen in figure \ref{fig:type}.
Here we can see the original concentration of sentences types in the testing dataset and the correctness of the classification following the mentioned method.

\begin{figure}[h]
\centering
\includegraphics[width=0.81\textwidth]{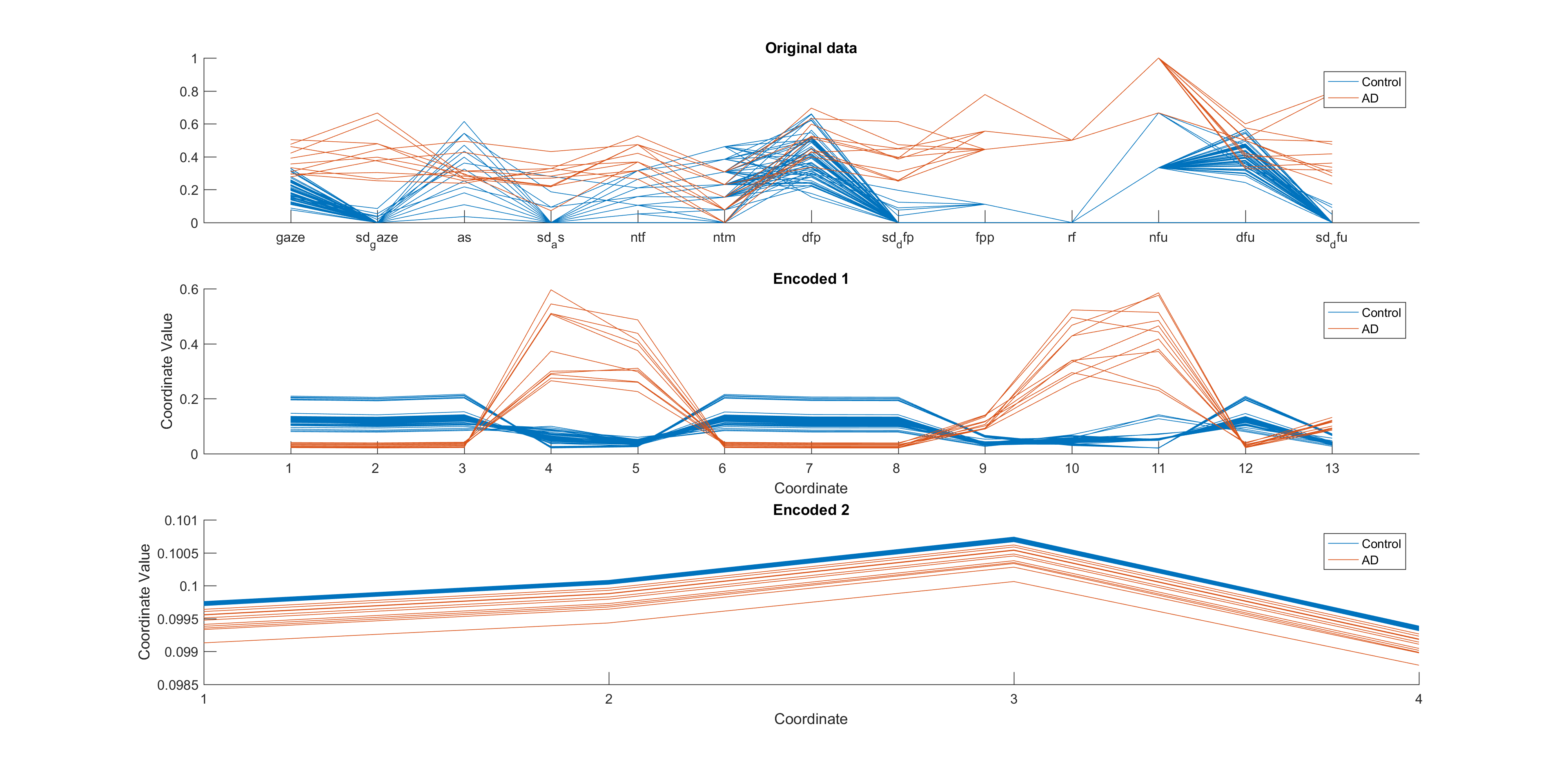}
\caption{Parallel coordinates plot with two subsets (one composed of \textit{AD} patients and the other of \textit{Control} subjects) of trials that have similar values for the input in each field and its codification during the different stages of the network. As expected, similar values encoded \say{together}. \textit{Control} subjects encoded closer than \textit{AD} subjects; This may be due the high \say{within group} variability of the \textit{AD} group.}
\label{fig:encoding-near}
\end{figure}

This result, added to the fact that neither was a concentration of misclassiﬁcation in a particular sentence, may suggest that most of the misclassiﬁcations occurred due to presumably stochastic processes.
In addition, the trained networks were evaluated using a spread result test in order to determine the softness of the model.
These tests checked if similar information is encoded in a similar way in the subsequent stages of the network.
A significant differentiation in later stages of encoding may show over-fitting in the network (and/or in the different stages).

Two subsets (one composed of \textit{AD} patients and the other of \textit{Control} subjects) of \textit{trials} that have similar values for the input in each field are shown in Figure \ref{fig:encoding-near}.
As shown, similar input values map to similar encoded values on each stage of the \textit{autoencoders}. This is because the modeled function is smooth. Furthermore, the data through the subsequent stages of codification tends to group.
These results have shown that the output information such as the encoding in the different stages are reliable.
On the other hand, they show that certain neurons in later stages tend to specialize on the detection of specific \textit{AD} or \textit{Control} input features.

\section{Conclussions and future work}
The results showed that using a deep-learning architecture for identifying the characteristic eye movements patterns of neurodegenerative diseases like Alzheimer's disease is a good approach since this technology is focused on pattern finding and is suitable for this work.
Moreover, the high performance in a per-trial classification approach, leads us to conclude that, since a single patient reads many sentences, the assertion rate per patient is higher than the $89.8\%$ accuracy reported in this work.
Using the policy that network outputs higher than $0.5$ are classified as \textit{AD} and below as \textit{Control}, if we tag each patient using \say{majority voting} over all of his/her read sentences, the network gets a $100\%$ classification accuracy for the testing set - $8$ well classified subjects from $8$ total -.
This was expected since, on this test set, the total number of missclassified sentences is $32$ and every patient, after the dataset cleaning, have $39.12$ ($SD=10.37$) mean sentences. 

\begin{table}[h!]
\begin{center}
\caption{Comparison of mean value given by the network and \say{severity of the disease} score given by psychiatrists.}{\begin{tabular}{lcccc}
\textbf{IDPat}& \textbf{Mean}& \textbf{S D} & \textbf{Score} & \textbf{Difference} \\ \cline{1-5}
58         & 0,97         & 0,17                     & 0,9   & 0,07       \\ 
57         & 0,95         & 0,17                     & 0,5   & 0,45       \\
66         & 0,49         & 0,32                     & 0,5   & 0,01       \\
60         & 0,95         & 0,16                     & 0,6   & 0,35       \\
56         & 0,96         & 0,16                     & 0,8   & 0,16       \\
55         & 0,94         & 0,17                     & 0,7   & 0,24       \\ 
63         & 0,87         & 0,25                     & 0,8   & 0,07       \\
64         & 0,51         & 0,34                     & 0,5   & 0,01       \\
70         & 0,90         & 0,24                     & 0,5   & 0,40       \\
69         & 0,84         & 0,25                     & 0,5   & 0,34       \\
65         & 0,91         & 0,22                     & 0,6   & 0,31       \\
59         & 0,58         & 0,36                     & 0,6   & 0,02       \\
71         & 0,75         & 0,33                     & 0,6   & 0,15       \\
62         & 0,76         & 0,32                     & 0,6   & 0,16       \\
67         & 0,47         & 0,35                     & 0,5   & 0,03       \\
68         & 0,40         & 0,31                     & 0,8   & 0,40       \\
53         & 0,84         & 0,31                     & 0,8   & 0,04       \\ \cline{1-5}
           &              &                          & \textbf{Mean}  & 0,19       \\ \cline{4-5}
           &              &                          & \textbf{SD}    & 0,15      
\end{tabular}}
\label{tab:score_comp}
\end{center}
\end{table}

Additionally, we asked the head psychiatrists of other \textit{AD} patients (that were not included in training process) to elaborate a score of the overall severity of the disease of each patient on the traditional tests using a scale from $0$ to $1$, without knowing the results given by our network.
The process of creating the score required that the physicians have a deep knowledge of the psychiatric history of each patient, the recompilation of the results of every neuropsychological tests made by the patient and its comprehension.
Table \ref{tab:score_comp} shows the scores given by the psychiatrists compared with the mean value obtained in our network for all the sentences read by the subjects and its standard deviation (S D).
As seen, for most of the patients, the values obtained are very similar to the scores given by the psychiatrists with a mean value of $0.19$ ($SD=0.15$). 
The obtained results show that the created marker is reasonably close to the score but involves a much simpler process.

Finding a better way to interpret the output of the classifier is left as future work. 
An improvement is required because values near $0.5$ are not determined to be identified as \textit{AD} or \textit{Control} (equal \say{probabilities}). 
The policy chosen in this work is a first rough approach that doesn't reflect the actual power of the network.
Using a fuzzy-logic encoder to obtain the overall diagnosis of a patient might lead to a more accurate result.
Determining whether the number given by the classiﬁer is related to the severity of the disease is left as a future improvement.
This task is particularly difficult since there are no ground truth measurements to corroborate the information due to the current psychological testing methods.
Anyway, although we adopted a per-trial classification approach, it's easy to think that the overall diagnosis may be related to a measurement extracted from the entire test and not from a single \textit{trial}.
As shown before, even with the policy used on this work and simply using the mean of the scores or the \say{controlling} label, this network behaved as expected.

\bibliographystyle{plainnat}
\bibliography{references}

\end{document}